\documentclass[letterpaper, twocolumn]{article} 
\usepackage{actiondiscovery}
\usepackage{times}  
\usepackage{helvet}  
\usepackage{courier}  
\usepackage[hyphens]{url}  
\usepackage{graphicx} 
\usepackage[numbers]{natbib}
\usepackage{caption}
\usepackage{algorithm}
\usepackage{algorithmic}

\usepackage{newfloat}
\usepackage{listings}
\DeclareCaptionStyle{ruled}{labelfont=normalfont,labelsep=colon,strut=off} 
\floatstyle{ruled}
\newfloat{listing}{tb}{lst}{}
\floatname{listing}{Listing}

\definecolor{refblue}{rgb}{0.21,0.49,0.74}
\usepackage[pagebackref,breaklinks,colorlinks,allcolors=refblue]{hyperref}
\usepackage[capitalize]{cleveref} 
\crefname{section}{Sec.}{Secs.}
\Crefname{section}{Section}{Sections}
\Crefname{table}{Table}{Tables}
\crefname{table}{Tab.}{Tabs.}

\title{Looking into the Unknown: Exploring Action Discovery for Segmentation of Known and Unknown Actions}
\author {
    Federico Spurio\textsuperscript{\rm 1,\rm 4},
    Emad Bahrami\textsuperscript{\rm 1,\rm 4},
    Olga Zatsarynna\textsuperscript{\rm 1,\rm4},
    Yazan Abu Farha\textsuperscript{\rm 2,\rm4},
    Gianpiero Francesca\textsuperscript{\rm 3},
    Juergen Gall\textsuperscript{\rm 1,\rm 4}
}
\affiliations {
    \textsuperscript{\rm 1}University of Bonn,
    \textsuperscript{\rm 2}Birzeit University,
    \textsuperscript{\rm 3}Toyota Motor Europe,
    \textsuperscript{\rm 4}Lamarr Institute for Machine Learning and Artificial Intelligence\\
}

\DeclareMathOperator{\argmin}{argmin}

\newcommand{\ie}{\textit{i.e.}\xspace}
\newcommand{\eg}{\textit{e.g.}\xspace}

\begin{document}

\maketitle
\thispagestyle{plain} 
\pagestyle{plain}

\begin{abstract}
We introduce {\it Action Discovery}, a novel setup within Temporal Action Segmentation that addresses the challenge of defining and annotating ambiguous actions and incomplete annotations in partially labeled datasets. In this setup, only a subset of actions - referred to as {\it known} actions - is annotated in the training data, while other {\it unknown} actions remain unlabeled. 
This scenario is particularly relevant in domains like neuroscience, where well-defined behaviors (e.g., walking, eating) coexist with subtle or infrequent actions that are often overlooked, as well as in applications where datasets are inherently partially annotated due to ambiguous or missing labels.
To address this problem, we propose a two-step approach that leverages the known annotations to guide both the temporal and semantic granularity of unknown action segments. First, we introduce the Granularity-Guided Segmentation Module (GGSM), which identifies temporal intervals for both known and unknown actions by mimicking the granularity of annotated actions. Second, we propose the Unknown Action Segment Assignment (UASA), which identifies semantically meaningful classes within the unknown actions, based on learned embedding similarities. 
We systematically explore the proposed setting of Action Discovery on three challenging datasets - Breakfast, 50Salads, and Desktop Assembly - demonstrating that our method considerably improves upon existing baselines.
\end{abstract}

\section{Introduction}
\label{sec:intro}

\begin{figure}[t]
    \centering
    \includegraphics[width=\columnwidth]{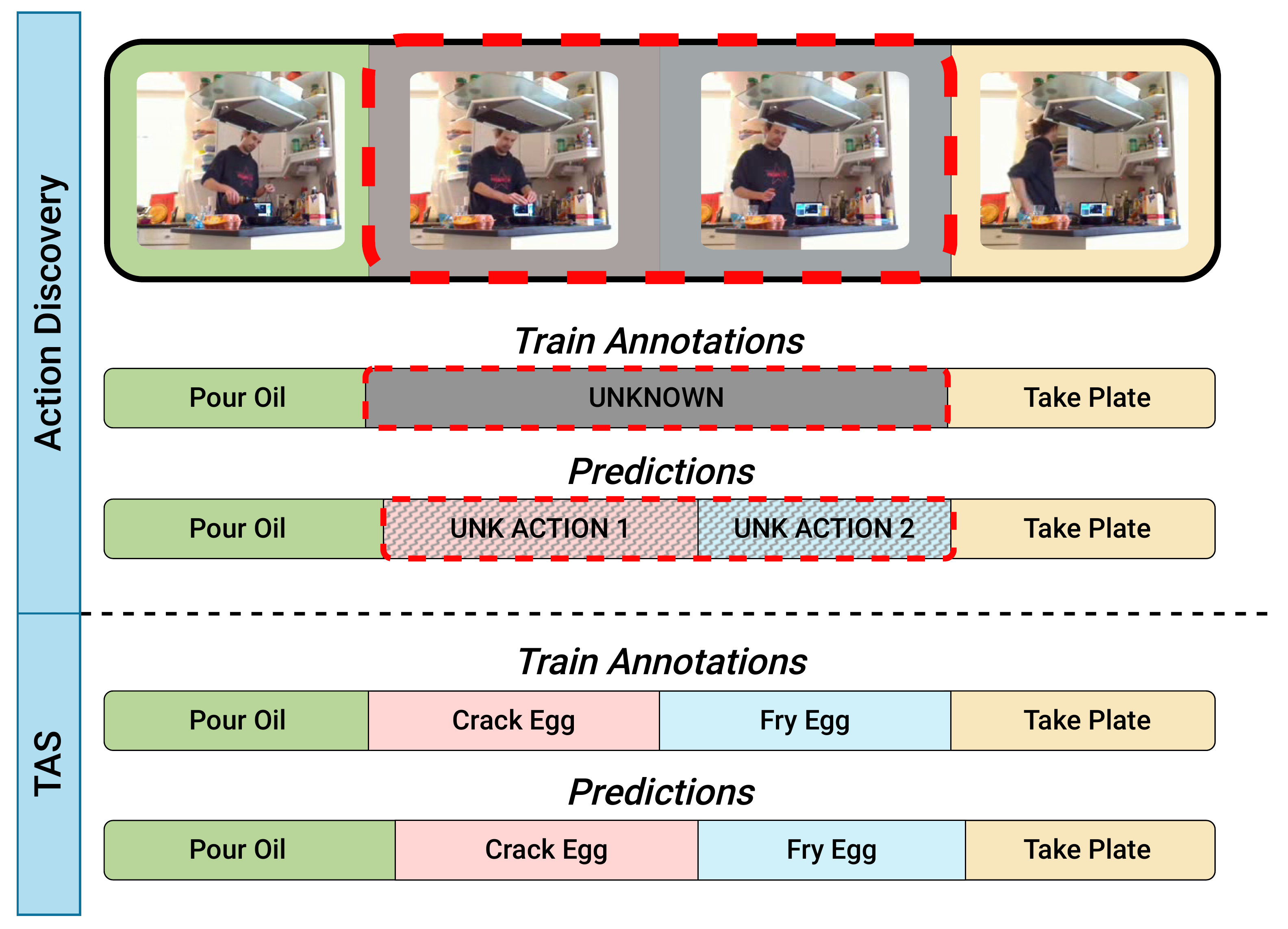}
    \captionsetup{type=figure}
    \caption{In standard Temporal Action Segmentation (TAS), all actions are assumed to be known, well-defined a priori, and annotated. In contrast, the proposed \textit{Action Discovery} protocol leverages only a subset of well-defined, known actions during training, enabling our approach to discover ambiguous or previously undefined unknown actions from partially annotated data.}
    \label{fig:teaser}
\end{figure}

Understanding and segmenting activities in videos has advanced significantly with fully-supervised methods~\cite{ltc2023bahrami, asformer, lu2024fact}, but these rely on costly, dense frame-level annotations. To reduce annotation effort, weaker forms of supervision, such as transcripts~\cite{chang2019d3tw, ding2018weakly, kuehne2017weakly, li2019weakly, richard2017weakly, richard2018neuralnetwork, lu2021weakly}, action sets~\cite{fayyaz2020sct, li2020set, richard2018action}, and timestamp labels~\cite{khan2022timestamp, li2021temporal, souri2022robust, rahaman2022generalized} have been explored.

However, these methods typically assume that all possible actions are known and annotations are complete — an assumption that rarely holds in real-world scenarios. Datasets are often partially annotated due to various factors, such as the prohibitive annotation effort, the presence of ambiguous actions, or instances where actions are summarily labeled as {\it ``background"} and require further refinement. Moreover, some actions may simply be ill-defined during the annotation phase. This situation is especially common in domains like animal and human behavior analysis, where even experts might overlook or be unaware of all occurring activities. Thus, discovering and modeling previously unlabeled actions is essential for a comprehensive understanding of the observed behaviors.

In the absence of a suitable benchmark to address this challenge, we introduce {\bf Action Discovery}, a novel setup within Temporal Action Segmentation (TAS), where, for the reasons mentioned above, only a subset of actions is labeled (see Figure~\ref{fig:teaser}). This framework helps to revisit partially annotated datasets and discover behaviors hidden within generic {\it background} labels. Known actions correspond to existing annotations, while ambiguous or omitted ones remain unlabeled and are treated as unknown.
We refer to the level of detail at which a known action is segmented as granularity. 
Unlike standard temporal action segmentation, this protocol focuses on identifying and segmenting unknown actions with the same granularity as known actions, while maintaining performance on known classes.

We propose the first method for this task. Given long, untrimmed videos, our approach outputs frame-wise labels for known actions and discovers new, previously unlabeled action segments. First, we introduce a {\bf Granularity-Guided Segmentation Module (GGSM)} to detect temporal spans of actions within unlabeled segments, allowing to detect new actions guided by the granularity of the known actions.
Furthermore, we propose an Unknown Action Segment Assignment (UASA) module to assign semantic labels to the discovered unknown actions, leveraging similarity in the embedding space without requiring the knowledge of the number of unknown classes.

To the best of our knowledge, this is the first attempt to address the Action Discovery protocol for Temporal Action Segmentation. 
To summarize, our key contributions are as follows:
\begin{itemize}
    \item We propose a challenging {\bf Action Discovery} setup that focuses on distinguishing between unlabeled actions and provide new dataset splits along with detailed statistics to support this task.
    \item We introduce a {\bf Granularity-Guided Segmentation Module (GGSM)} to accurately identify temporal action spans within unlabeled segments, based on the granularity of known annotations. 
    \item  By leveraging coarse-level annotations, our method can segment and identify fine-grained actions within unlabeled segments, addressing the challenge of unlabeled actions in training data.
    \item Our similarity-based approach, {\bf Unknown Action Segment Assignment (UASA)}, assigns the most plausible labels to each detected segment within unknown actions. This is achieved by leveraging embeddings from the action segmentation backbone, without requiring the knowledge of the number of unknown classes.
\end{itemize}

\section{Related works}
\label{sec:related}

\paragraph{Fully-Supervised Action Segmentation.} Early methods tackled action segmentation with HMMs~\cite{kuehne2016end, tang2012latent} or sliding windows~\cite{karaman2014, rohrbach2012}. Later, sequence modeling via length and language models~\cite{richard2016temporal} and frame-based approaches like MS-TCN~\cite{farha2019ms, li2020ms} advanced performance, but struggled with long-range dependencies. Graph-based reasoning~\cite{huang2020gbtr} and boundary regression~\cite{ishikawa2021dab} improved localization, while ASFormer~\cite{asformer} leveraged transformers but faced over-segmentation. UVAST~\cite{behrmann2022unified} mitigated this with segment-level cues, and LTContext~\cite{ltc2023bahrami} introduced sparse attention. More recently, DiffAct~\cite{liu2023diffusion} applied diffusion models, and FACT~\cite{lu2024fact} incorporated textual transcripts. 

\paragraph{Weakly- and Semi-Supervised Action Segmentation.} To reduce annotation costs, weakly-supervised methods have explored four supervision types: transcripts~\cite{chang2019d3tw, ding2018weakly, kuehne2017weakly, li2019weakly, richard2017weakly, richard2018neuralnetwork, lu2021weakly}, using ordered lists of actions; action sets~\cite{fayyaz2020sct, li2020set, richard2018action}, relying on unordered lists of present actions; timestamps~\cite{khan2022timestamp, li2021temporal}, where only one frame per segment is annotated; and sparse timestamps~\cite{souri2022robust, rahaman2022generalized}, which allow sporadic missing annotations. UVAST~\cite{behrmann2022unified} unified these supervision types with fully-supervised learning. Semi-supervised approaches~\cite{singhania2022icc, ding2022sslosses} further reduced annotation requirements by training on partially annotated datasets, where only a subset of videos is fully labeled while the rest remain unlabeled.
In contrast, we address scenarios where some action classes remain completely unlabeled during training, requiring the model to discover and differentiate unknown actions without explicit supervision. 
Although these classes are unlabeled, the model is exposed to them during training, making our setting fundamentally different from the Open Set scenario. 
While Open Set Action Localization has been studied~\cite{bao2022opental, zhang2023owtal, gupta2024open, bao2025exploiting}, Open Set Action Segmentation remains largely unexplored. 
Unlike localization, which detects action intervals, segmentation requires dense, frame-level predictions. Our framework further differs by requiring the model to distinguish among multiple unknown actions, beyond a binary seen/unseen setting.

\paragraph{Unsupervised Action Segmentation.} Unsupervised TAS has been addressed at the activity-level~\cite{kukleva2019unsupervised, kumar2022tot, xu2024asot, spurio2025hvq} and video-level~\cite{sarfraz2021twfinch}. However, activity-level methods struggle with fine-grained classes, while video-level methods lack generalization. Our approach leverages known annotations to segment unknown actions across videos, without relying on limiting the number of unknown actions a priori.

\section{Action Discovery}
\label{sec:method}

\begin{figure*}[tbp]
    \centering
    \includegraphics[width=0.75\linewidth]{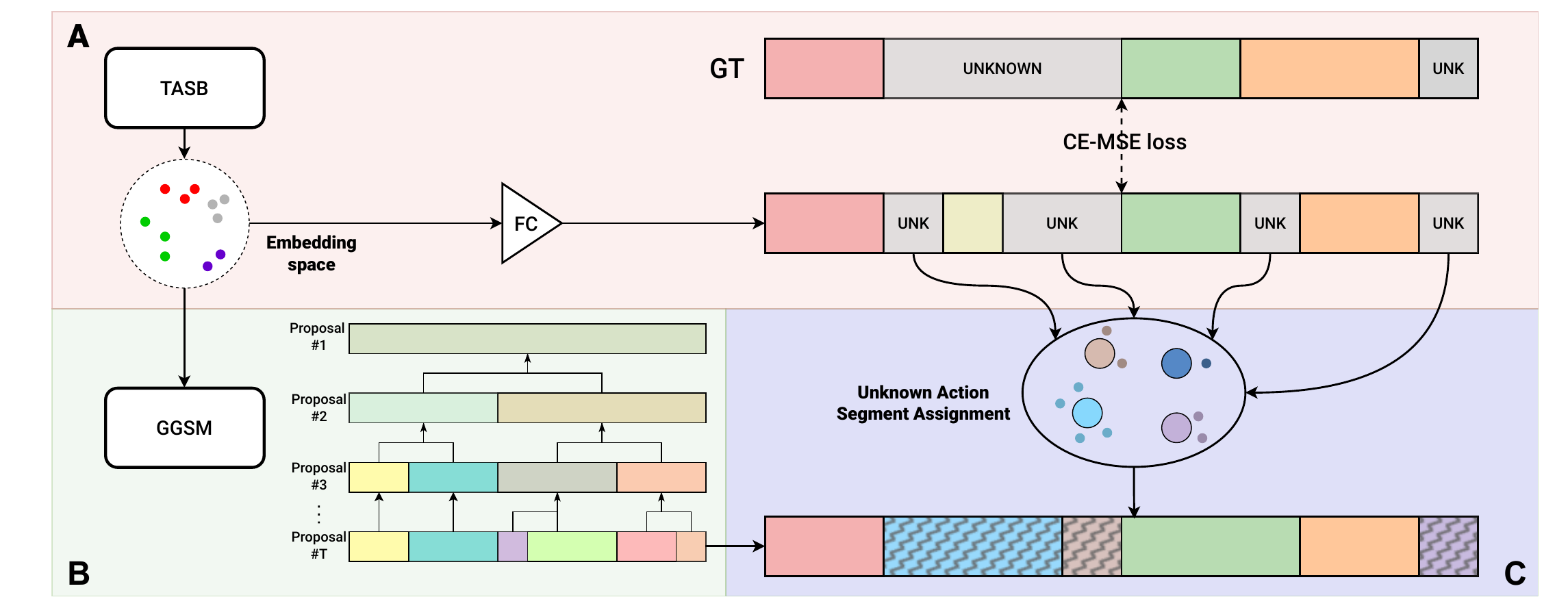}
    \caption{Overview of our architecture. (A) TASB is trained to discriminate between known and unknown actions. (B) The GGSM clusters the learned embeddings hierarchically, until it reaches the comparable granularity with the ground-truth. (C) In parallel, UASA assigns semantic labels to unknown segments. Final predictions are obtained by majority voting within GGSM intervals.}
    \label{fig:model}
\end{figure*}

Temporal action segmentation involves predicting an action label for each frame of a video. Current models assume that all the actions are known and annotated during training. However, in many scenarios not all the actions are known a priori during the annotation, either because these actions are not of interest at the time of annotation or due to changes in annotation protocol. 

To address this limitation, we introduce a novel setting for action segmentation called \textit{Action Discovery}. In this setting, actions $A$ are divided into two sets: known $\phi_{act} \subset A$ and unknown $\phi_{unk} \subset A$, such that $\phi_{act} \cap \phi_{unk}=\emptyset$ and $\phi_{act} \cup \phi_{unk}=A$. 
Information about unknown actions is unavailable, and all unknown actions are considered as background, \ie, they are not annotated. The challenge of Action Discovery lies in segmenting the unknown actions and classifying them into semantically meaningful actions across videos. While the known actions are annotated in the training data, neither known nor unknown actions are annotated in the test data.
The overview of our architecture, which learns from known actions to discover further unknown actions in the training data and temporally segments both in the test data, is shown in Figure \ref{fig:model}.

First, we treat unknown action segments as an additional label and train an action segmentation model to classify each video into segments of known actions, along with a single combined segment for unknown actions. Then, since a predicted unknown segment may contain multiple distinct action segments, we leverage embeddings from the trained model as input to our proposed Granularity-Guided Segmentation Module (GGSM). This module refines detected unknown segments into action-specific unknown segments.

To assign labels to unknown action segments, we employ a similarity-based method that groups frames with similar features. Using unknown embeddings from a Temporal Action Segmentation Backbone, we first estimate $K$, the number of unknown actions, via Gaussian Mixture Models. We then apply K-means clustering to assign action labels to frames based on their proximity in the embedding space. 

\paragraph{Temporal Action Segmentation Backbone (TASB)} During the training, the ground-truth labels are composed of {\it known actions}, with labels $[c_1,...,c_{|\phi_{act}|}] \in \phi_{act}$, and {\it unknown actions}, labeled by {\tt UNK}. The Temporal Action Segmentation Backbone (TASB) is trained to predict $|\phi_{act}|+1$ number of classes, where the additional class is for {\it unknown actions}. We call $E=[e_1,...,e_T] \in \mathbb{R}^{T \times D}$ the embeddings learned by the network before the classification layer, where $D$ is the embedding dimension.
The backbone is trained using a combination of Cross Entropy loss and truncated Mean Squared Error loss ($\mathcal{L}_{backbone}$) as in \cite{farha2019ms}. To prevent the collapse of unknown embeddings, 
we incorporate a contrastive loss specifically for unknown actions. Within each unknown segment, we encourage temporally close frames to stay close in the embedding space while pushing apart temporally distant frames. We achieve this using the contrastive loss \cite{oord2018representation}, where positive pairs are sampled from a truncated Gaussian distribution centered on the anchor frame, while negative pairs are drawn from frames outside this window. The contrastive loss is then computed by
\begin{align*}
\mathcal{L}_{contr} &= -\frac{1}{T}\sum_{t=1}^{T}\log\frac{\exp(\text{sim}(e_t, e_{\text{pos}(t)})/\tau)}{\sum_{k \in \text{neg(t)} \cup \text{pos(t)}}\exp(\text{sim}(e_t, e_k)/\tau)} \\
\end{align*}
where $e_{pos(t)}$ and $e_{neg(t)}$ represent correspondingly the positive and negative samples for $e_t$, and sim is the cosine similarity. We set the temperature $\tau$ to $0.4$.

The final loss is then defined as the weighted sum between backbone and contrastive loss: \(\mathcal{L}_{final} = \mathcal{L}_{backbone} + \lambda \mathcal{L}_{contr}\).

\paragraph{Granularity-Guided Segmentation Module (GGSM)}
\label{sec:hbdm}
To properly segment unknown actions, \ie find boundaries for every unknown action, we propose a Granularity-Guided Segmentation Module (GGSM), capable of segmenting the unknown actions based on the granularity of the known action segments. Our approach relies on the temporal intervals identified by annotators as behaviors or actions, unlike unsupervised boundary detectors~\cite{sarfraz2021twfinch, Du2022ABD, mounir23streamer}, which depend solely on the inherent properties of the data. Given the same video with two different levels of annotation granularity, unsupervised methods would produce the same segmentation, whereas our approach would not. To demonstrate the superiority of our approach, we adapt one of the these methods and evaluate its performance in Section~\ref{sec:twfinch}.   

In particular, GGSM takes as input the representation embeddings $E \in \mathbb{R}^{T\times D}$ learned by the Temporal Action Segmentation Backbone for the entire video and clusters them hierarchically. 
GGSM begins with frame-level embeddings and progressively merges clusters to achieve a segmentation of both known and unknown actions. Specifically, each frame initially forms its own cluster, and over multiple iterations, clusters are merged based on a distance metric. At each step, the two clusters that are closest to each other are merged into a single cluster. This process continues until all data points belong to the same cluster. 
The algorithm requires a distance metric to measure the distance between data points or clusters. In this case, we use the Euclidean distance. The distance between two clusters is defined as the maximum distance between any two points, one from each cluster. This specific approach is often referred to as the Farthest Point Algorithm. Formally, the distance between two clusters $\mathcal{V}$ and $\mathcal{S}$ is defined as follows:
\begin{equation}    
\begin{aligned}
\label{eq:ward}
D(\mathcal{S}, \mathcal{V}) = \max_{\mathbf{s} \in \mathcal{S}, \mathbf{q} \in \mathcal{V}}  d(\mathbf{s}, \mathbf{q}) 
\end{aligned}
\end{equation}
where $d(\cdot, \cdot)$ is the Euclidean distance and $\mathbf{s}$ and $\mathbf{q}$ are two points from cluster $\mathcal{S}$ and $\mathcal{V}$, respectively. 

The clustering algorithm generates multiple possible segmentations of the input video at each level of the hierarchy, which we refer to as proposals. To select the best proposal from GGSM, we compute a score based on the alignment between the proposed temporal intervals and the temporal spans of known actions. During training, we use the ground-truth temporal spans of known actions for this alignment; during inference, we rely on the predicted temporal spans of known actions instead. This score, independent of action labels, is calculated as the Intersection over Union (IoU) between the $M$ proposals and $N$ known actions segments:

\begin{equation}  
    \label{eq:iou1d}
    IoU = \frac{1}{MN} \sum_{i=1}^{M} \sum_{j=1}^{N} \text{IoU}(I_i^{p}, I_{j}^{gt}).
\end{equation}

We add to the computation of the IoU a balancing term that prefers segments of the same length if two segments overlap but are not perfectly aligned. Using $s_i$ and $f_i$ as the start and end frame of a segment $I_i$, the unbalanced score ($\text{IoU}_{unbal}$) is computed as
\begin{equation}
    \text{IoU}_{unbal}(I_i, I_j) = \frac{\max(0, \min(f_i, f_j) - \max(s_i, s_j))}{\max(f_i, f_j) - \min(s_i, s_j)},
\end{equation}
and we multiply it by the balancing term
\begin{equation}
    \text{IoU}(I_i, I_j) = \text{IoU}_{unbal} \times e^{-\alpha |(f_i-s_i) - (f_j-s_j)|}
\end{equation}

where $|(f_1-s_1) - (f_2-s_2)|$ is the absolute length difference of the two segments.

We finally take the proposal with the highest IoU. This proposal defines the granularity of the unknown actions, given the context and granularity of the known actions. In contrast to unsupervised approaches where the number of unknown classes needs to be provided, our approach estimates the granularity and thus the number of unknown actions directly from the data.    

\paragraph{Unknown Action Segment Assignment (UASA)}
To classify unknown actions while ensuring semantic consistency across videos, we utilize the temporal intervals predicted by the Granularity-Guided Segmentation Module (GGSM) to extract unknown segments and cluster their mean representations. 

For each video, we first identify unknown segments based on the temporal spans predicted by GGSM. For each segment $S_k$, we then average its frame embeddings \(e_i\)
\begin{equation} 
\mu_k = \frac{1}{|S_k|} \sum_{e_i \in S_k} e_i. 
\end{equation}
Unlike unsupervised action segmentation, we do not predefine the parameter $K$, which represents the maximum number of unknown classes. Instead, we initialize multiple Gaussian Mixture Models (GMMs) with increasing values of $K$ and select the one that yields the lowest Bayesian Information Criterion (BIC)~\cite{Schwarz1978BIC} score.
Once the mean representations are obtained, we perform K-Means clustering over $\{\mu_k\}$. The resulting centroids are then used for classifying the unknown actions. Each unknown segment is assigned the label of the closest centroid $z_j$, determined via Euclidean distance: \(z_k^{closest} = \argmin_{j} (d(\mu_k, z_j))\),
where $d(\mu_k,z_j)$ is the Euclidean distance between the segment mean and centroid $z_j$. This process ensures that unknown actions are assigned coherent labels based on segment-level information rather than frame-wise clustering.

\paragraph{Inference}
During inference, our method does not rely on any annotations, whether related to known or unknown actions. The backbone, TASB, first produces preliminary predictions that distinguish between specific known classes and a general unknown class. Next, GGSM refines the segmentation by proposing a new division based on the predicted known actions. Finally, UASA classifies each unknown segment identified by GGSM into one of the $K$ unknown classes, where $K$ is determined on the training set.

\section{Experiments}

\subsection{Datasets, Evaluation and Metrics}
To ensure a meaningful and reproducible evaluation, and for consistency with prior benchmarks, we conduct experiments on three widely-used datasets for Temporal Action Segmentation: {\bf Breakfast}~\cite{kuehne2014breakfast}, {\bf 50Salads}~\cite{stein2013fs}, and {\bf DesktopAssembly}~\cite{kumar2022tot}. 

For these existing datasets, we design a partition of actions into known and unknown, to simulate the challenges of Action Discovery and partially annotated data. The resulting partitions and corresponding statistics are summarized in Table~\ref{tab:statistics}.

{\bf Breakfast} contains 1,712 videos (70 hours) covering 48 actions from 10 kitchen activities. We follow the 4 official train-test splits~\cite{kuehne2014breakfast}, designating 33 actions as known and 15 as unknown.
{\bf 50Salads} includes 50 videos (5.5 hours) of salad preparation, evaluated with five-fold cross-validation~\cite{farha2019ms, asformer, behrmann2022unified}. We consider two annotation granularities: {\bf Fine} (19 actions: 13 known, 6 unknown) and {\bf Coarse} (11 actions: 7 known, 4 unknown), where semantically similar actions are merged.
{\bf DesktopAssembly} consists of 128 videos (3 hours) with 22 desktop assembly actions, and 65\% of unknown segments involve multiple actions (16 known actions and 6 unknown actions).

For features, we use I3D~\cite{carreira2017i3d} for Breakfast and 50Salads, and dataset-specific features from~\cite{kumar2022tot} for DesktopAssembly, following the train-test split from~\cite{tran2024ufsa}.

\begin{table}[tbp]    
    \setlength{\tabcolsep}{4.pt}
    \centering
    \resizebox{\columnwidth}{!}{%
    \begin{tabular}{|l|c|c|c|c|}
        \hline
        \textbf{Statistics} & \textbf{Breakfast} & \makecell[c]{{\bf 50Salads} \\{\bf Fine}} & \makecell[c]{{\bf 50Salads}\\{\bf Coarse}} & {\bf DA} \\
        \hline
        \multicolumn{5}{|c|}{Segment statistics \textit{(\%)}} \\
        \hline
        \rowcolor{teal!25} Unk. segs. ($>1$ unk. action) & $48.5$ & $33.3$ & $42.9$ & $65.5$ \\ 
        \rowcolor{teal!25} Unk. segs. ($>2$ unk. actions) & $11.4$ & $14.1$ & $25.4$ & $33.2$ \\ 
        \hline
        \multicolumn{5}{|c|}{Segment statistics \textit{(num.)}} \\
        \hline
        \rowcolor{blue!15} Total seg. & $10,458$ & $889$ & $496$ & $2,582$ \\
        \rowcolor{blue!15} Total unknown seg. & $1,927$ & $213$ & $126$ & $245$ \\
        \hline
        \multicolumn{5}{|c|}{Action statistics \textit{(num.)}} \\
        \hline
        \rowcolor{orange!15} Known actions & $33$ & $13$ & $7$ & $16$ \\
        \rowcolor{orange!15} Unknown actions & $15$ & $6$ & $4$ & $6$ \\
        \hline
        \multicolumn{5}{|c|}{Frame statistics} \\
        \hline
        \rowcolor{purple!15} Known frames \textit{(num.)} & $2,716,905$ & $420,578$ & $398,942$ & $63,966$ \\
        \rowcolor{purple!15} Unknown frames \textit{(num.)} & $873,994$ & $157,017$ & $178,653$ & $26,506$ \\
        \rowcolor{purple!15} Unknown frames \textit{(\%)} & $24.3$ & $27.2$ & $30.9$ & $29.3$ \\
        \hline
    \end{tabular}
    }
    \captionsetup{type=table}
    \caption{Statistics of the proposed Action Discovery splits for the three datasets, detailing the distribution of known and unknown actions. DA denotes DesktopAssembly.}
    \label{tab:statistics}
\end{table}

\paragraph{Evaluation and Metrics}
We evaluate known and unknown actions separately. While the primary focus is on evaluating the unknown actions, we also report results for known actions to ensure that performance on them does not degrade significantly while improving unknown action segmentation.
We assess our proposed protocol using the same metrics as Fully-Supervised Temporal Action Segmentation. Specifically, we use Mean over Frame (MoF), segmental Edit distance, and the segmental F1 score at overlapping thresholds of $10\%$, $25\%$ and $50\%$, denoted as $F1@\{10,25,50\}$. The Intersection over Union (IoU) ratio serves as the overlapping threshold. 
Since Action Discovery focuses on unknown action segmentation, we introduce a masked version of these metrics to evaluate known and unknown actions separately. To achieve this, we mask unknown actions when evaluating known actions and vice versa.

\paragraph{Unknown Evaluation} In order to measure quantitatively the performance of our proposed architecture on unknown actions, we follow the standard approach used in Unsupervised Action Segmentation~\cite{alayrac2016narrated, sener2018unsupervised, kukleva2019unsupervised}. This consists in aligning the predicted unknown classes with the ground-truth classes thanks to the one-to-one matching performed by the Hungarian algorithm. After the matching, every unknown cluster centroids are aligned with only one semantic label among the ground-truth classes. It is important to note that the mapping is only used for evaluation. The model does not have access to ground-truth of unknowns neither during training nor during inference. 
Differently from the unsupervised Action Segmentation protocol, the Hungarian matching is done at dataset level, rather than activity level, and we do not provide the parameter $K$, which represents the maximum number of actions in the dataset. Instead, we allow the network to estimate this value autonomously.  
In a real-world application of Action Discovery, it is up to experts to assess the quality of the segmentation. Furthermore, we report the results on the test data, \ie, the model does not have access to any annotations of the test sequences.   
%
\subsection{Implementation Details}
As temporal action segmentation backbone, we use MS-TCN with the same parameter configuration as in \cite{farha2019ms}. 
The sets for known and unknown actions are carefully selected to create a challenging setup. As shown in Table \ref{tab:statistics}, we ensure that the majority of unknown segments do not contain only a single action -- which would be trivial for the network to distinguish. Instead, in all datasets, at least $30\%$ of the unknown segments consist of multiple actions, making the segmentation task more demanding. 
The list of known and unknown actions for every dataset is provided in the supplementary material.

\subsection{Comparison with the Baselines}
In this section, we compare the proposed approach for Action Discovery against several baselines. All methods use MS-TCN \cite{farha2019ms} as backbone, differing only in how they handle unknown actions during training and evaluation. {\bf Baseline} is trained to distinguish known actions and a single generic class for all unknown actions, without differentiating among them. During evaluation, all unknown actions are assigned the same label, matched with the most frequent unknown action via Hungarian Matching. {\bf Baseline+GGSM} incorporates the Granularity-Guided Segmentation Module (GGSM) to refine the segmentation of unknown segments and reduce over-segmentation. However, as in Baseline, all unknown frames are still assigned a single label. {\bf Baseline+UASA} adds the Unknown Action Segment Assignment (UASA) module to cluster unknown actions based on embedding space similarity. In this setting, GGSM is disabled to isolate UASA's effect.

We present the results for all datasets in Table~\ref{tab:segm-result}. The table also includes the performance of {\bf Fully-Supervised MS-TCN}, trained under the standard fully-supervised temporal action segmentation protocol, \ie without distinguishing between known and unknown, with all action labels available during training.

\begin{table*}[tbp]
    \centering
    \begin{minipage}{0.48\textwidth}
        \centering
        \renewcommand{\arraystretch}{1.2}  
        \resizebox{\textwidth}{!}{%
        \begin{tabular}{|c|ccccc|ccccc|}
            \hline
            \multirow{2}{*}[-0.2em]{\bf Model}  & \multicolumn{5}{c|}{\bf Known} 
            & \multicolumn{5}{c|}{\bf Unknown} \\[0.5ex]
            \cline{2-11}
            & {\it MoF} & {\it Edit} & \multicolumn{3}{c|}{\it F1@\{10,25,50\}} 
            & {\it MoF} & {\it Edit} & \multicolumn{3}{c|}{\it F1@\{10,25,50\}} \\[0.5ex]
            \hline
            \rowcolor{pink!30} \multicolumn{11}{|c|}{\bf Breakfast} \\
            \hline
            \rowcolor{gray!15} F.-S. MS-TCN* &
            \textit{67.4} & \textit{35.7} & \textit{48.4} & \textit{44.6} & \textit{35.3} &
            \textit{61.8} & \textit{55.3} & \textit{66.3} & \textit{63.0} & \textit{54.7} \\
            \hline
            Baseline & 
            {\bf 60.2} & {\bf 65.9} & {\bf 70.6} & {\bf 64.9} & {\bf 50.0} &
            10.5 & 8.8 & 10.7 & 10.1 & 8.7 \\
            B. + GGSM & 
            48.5 & 55.2 & 43.4 & 40.7 & 32.0 &
            11.0 & 11.2 & 13.2 & 12.1 & 9.3 \\
            \cline{2-11}
            B. + UASA & 
            {\bf 60.2} & {\bf 65.9} & {\bf 70.6} & {\bf 64.9} & {\bf 50.0} &
            16.6 & 20.4 & 23.1 & 17.7 & 8.6 \\
            Ours (GGSM + UASA) & 
            56.0 & 54.8 & 56.2 & 52.5 & 41.3 &
            {\bf 23.5} & {\bf 27.1} & {\bf 27.7} & {\bf 25.6} & {\bf 18.6} \\
    
            \hline
            \rowcolor{pink!30} \multicolumn{11}{|c|}{\bf DesktopAssembly} \\
            \hline
            
            \rowcolor{gray!15} F.-S. MS-TCN* &
            \textit{93.2} & \textit{97.3} & \textit{98.5} & \textit{98.5} & \textit{96.7} &
            \textit{91.9} & \textit{95.4} & \textit{97.7} & \textit{97.0} & \textit{94.4} \\
        
            \hline
            Baseline &         
            {\bf 83.8} & {\bf 96.3} & {\bf 96.2} & {\bf 94.0} & {\bf 85.9} &
            39.9 & 17.4 & 25.6 & 25.6 & 14.2 \\
            B. + GGSM & 
            80.4 & 88.1 & 87.9 & 85.5 & 82.3 &
            39.2 & 17.4 & 25.8 & 25.8 & 14.3 \\
            \cline{2-11}
            B. + UASA & 
            {\bf 83.8} & {\bf 96.3} & {\bf 96.2} & {\bf 94.0} & {\bf 85.9} &
            50.1 & 42.5 & 61.6 & 54.6 & 29.3 \\
            Ours (GGSM + UASA) & 
            77.5 & 86.8 & 90.1 & 86.6 & 81.1 &
            {\bf 51.3} & {\bf 52.8} & {\bf 62.5} & {\bf 58.0} & {\bf 45.4} \\
            
            \hline
        \end{tabular}%
        }

    \end{minipage}
    \hfill
    \begin{minipage}{0.48\textwidth}
        \centering
        \renewcommand{\arraystretch}{1.2}  
        \resizebox{\textwidth}{!}{%
        \begin{tabular}{|c|ccccc|ccccc|}
            \hline
            \multirow{2}{*}[-0.2em]{\bf Model}  & \multicolumn{5}{c|}{\bf Known} 
            & \multicolumn{5}{c|}{\bf Unknown} \\[0.5ex]
            \cline{2-11}
            & {\it MoF} & {\it Edit} & \multicolumn{3}{c|}{\it F1@\{10,25,50\}} 
            & {\it MoF} & {\it Edit} & \multicolumn{3}{c|}{\it F1@\{10,25,50\}} \\[0.5ex]            
            
            \hline 
            \rowcolor{pink!30} \multicolumn{11}{|c|}{\bf 50Salads Fine} \\
            \hline       
    
            \rowcolor{gray!15} F.-S. MS-TCN* &
            \textit{80.8} & \textit{68.6} & \textit{75.8} & \textit{74.0} & \textit{67.2} &
            \textit{70.6} & \textit{69.1} & \textit{78.1} & \textit{75.3} & \textit{67.2} \\
        
            \hline
            Baseline & 
            {\bf 71.8} & {\bf 67.2} & {\bf 71.6} & 68.0 & 60.2 &
            24.6 & 16.9 & 20.3 & 19.3 & 12.3 \\
            B. + GGSM & 
            68.1 & 61.2 & 66.5 & 65.4 & 59.8 &
            25.3 & 16.9 & 21.8 & 21.2 & 13.1 \\
            \cline{2-11}
            B. + UASA & 
            {\bf 71.8} & {\bf 67.2} & {\bf 71.6} & 68.0 & 60.2 &
            34.1 & 38.1 & 39.9 & 34.3 & 23.1 \\
            Ours (GGSM + UASA) & 
            69.4 & 62.9 & 69.6 & {\bf 68.5} & {\bf 62.4} &
            {\bf 37.2} & {\bf 40.1} & {\bf 45.1} & {\bf 41.5} & {\bf 27.1} \\
            
            \hline
            \rowcolor{pink!30} \multicolumn{11}{|c|}{\bf 50 Salads Coarse} \\
            \hline
    
            \rowcolor{gray!15} F.-S. MS-TCN* &
            \textit{78.4} & \textit{30.1} & \textit{42.6} & \textit{41.4} & \textit{37.6} &
            \textit{78.5} & \textit{62.1} & \textit{71.7} & \textit{70.5} & \textit{64.8} \\
        
            \hline
            Baseline & 
            {\bf 78.0} & 73.3 & 80.3 & 78.1 & 73.8 &
            41.5 & 25.7 & 33.8 & 33.3 & 18.4 \\
            B. + GGSM & 
            72.5 & {\bf 75.1} & {\bf 82.5} & {\bf 81.1} & {\bf 75.3} &
            40.3 & 25.6 & 34.8 & 34.3 & 19.1 \\
            \cline{2-11}
            B. + UASA & 
            {\bf 78.0} & 73.3 & 80.3 & 78.1 & 73.8 &
            33.8 & 39.3 & 37.6 & 33.7 & 16.2 \\
            Ours (GGSM + UASA) & 
            72.4 & 72.1 & 80.4 & 78.4 & 71.3 &
            {\bf 46.4} & {\bf 45.6} & {\bf 49.3} & {\bf 46.5} & {\bf 36.6} \\
            
            \hline
        \end{tabular}%
        }
    \end{minipage}
    \caption{Segmentation results on the three datasets proposed: {\bf Breakfast}, {\bf 50Salads (Fine and Coarse)} and {\bf DesktopAssembly}. Best results are in {\bf bold}. * indicates methods trained with all the labels, without distinction between known and unknown. `B.' denotes the Baseline and `F.-S.' stands for Fully-Supervised.}
    \label{tab:segm-result}
\end{table*}

\paragraph{Breakfast} 
\label{sec:exp-break}
The Breakfast dataset shows that the Baseline model performs some segmentation on unknown actions, but struggles to distinguish between them, assigning all unknown frames to a single generic class. Despite this, it outperforms the fully-supervised version on known actions, as it focuses on a smaller set of action classes. Adding the proposed GGSM improves segmentation for unknown actions by addressing over-segmentation and producing more coherent segments, but at the cost of performance on known actions. However, without UASA, unknown actions remain under-classified, as the improvement is solely due to reducing over-segmentation. While UASA alone does not improve performance on unknown actions, its combination with GGSM significantly boosts the metrics, primarily due to the proper initialization of UASA. Importantly, only GGSM impacts performance on known actions.

\paragraph{50Salads} In the 50Salads dataset, the results are similar to those observed on Breakfast, with UASA improving performance on unknown actions, particularly in the Fine split. As with Breakfast, combining UASA with GGSM yields the best results across both Fine and Coarse splits, highlighting the importance of segment initialization in UASA.
Analyzing the results across Fine and Coarse splits, we observe that our method successfully adapts to different levels of granularity in the annotations of the same dataset.

\paragraph{DesktopAssembly} DesktopAssembly is the easiest of the three datasets, as the action order is nearly fixed across videos, which is reflected in the high performance of the fully-supervised MS-TCN. However, due to the high proportion (65\%) of unknown segments containing multiple actions, the Baseline’s performance drops significantly. In this challenging setting, Baseline+UASA achieves good segmentation of unknown actions due to the well-structured embedding space, but the best performance is still achieved by combining UASA with GGSM, as with the other two datasets.

\subsection{Ablations}
\paragraph{Ablation on Handling Unknown}
As an alternative, unsupervised action segmentation approaches can be applied to the embeddings labeled as unknown by the backbone.
To assess this, we evaluate ASOT~\cite{xu2024asot}, a state-of-the-art unsupervised action segmentation method, on these embeddings. As shown in Table \ref{tab:asot}, our approach outperforms ASOT by a large margin, demonstrating the effectiveness of GGSM in refining unknown segments and improving overall segmentation quality.

\paragraph{Ablation on Unknown Segmentation}
\label{sec:twfinch}
In Section \ref{sec:method}, we stated that our method can segment unknown actions based on the granularity of the known actions. To validate this, we compare our approach with TW-FINCH~\cite{sarfraz2021twfinch}, a state-of-the-art unsupervised method for detecting boundaries in videos.  
The implementation of TW-FINCH requires specifying the number of actions $K$, set as the average number of actions per activity. However, using this value will introduce prior knowledge about the number of unknown actions, which is not allowed in our setting. Instead, we set $K$ as the number of known actions per video plus $2$, striking a balance to account for possible unknown actions. 
As the results in Table~\ref{tab:finch} demonstrate, leveraging the granularity of known actions significantly improves the segmentation quality of unknown actions.   

\begin{table}[tbp]
    \centering
    \centering
    \setlength{\tabcolsep}{4.8pt}
    \renewcommand{\arraystretch}{1.3}  
    \resizebox{\columnwidth}{!}{%
    \begin{tabular}{|c|ccccc|ccccc|}
        \Xcline{1-11}{0.6pt}
        \multicolumn{11}{|c|}{\bf 50Salads Fine} \\
        \Xcline{1-11}{0.6pt}
        \multirow{2}{*}[-0.2em]{\bf Model} & \multicolumn{5}{c|}{\bf Known} 
        & \multicolumn{5}{c|}{\bf Unknown} \\[0.5ex]
        \cline{2-11}
        & {\it MoF} & {\it Edit} & \multicolumn{3}{c|}{\it F1@\{10,25,50\}} 
        & {\it MoF} & {\it Edit} & \multicolumn{3}{c|}{\it F1@\{10,25,50\}} \\[0.5ex]
        \hline
        \rowcolor{orange!15} Ours (GGSM + UASA) &
        69.4 & 62.9 & 69.6 & {\bf 68.5} & {\bf 62.4} &
        {\bf 37.2} & {\bf 40.1} & {\bf 45.1} & {\bf 41.5} & {\bf 27.1} \\
        B. + ASOT~\cite{xu2024asot} & 
        {\bf 71.8} & {\bf 67.2} & {\bf 71.6} & 68.0 & 60.2 &
        29.8 & 26.2 & 22.6 & 19.3 & 15.2 \\
        \hline
    \end{tabular}%
    }
    \caption{Performance comparison between our approach and ASOT~\cite{xu2024asot}, a state-of-the-art unsupervised action segmentation method applied to embeddings labeled as unknown by TASB. Best results are in {\bf bold}. `B.' denotes the Baseline.}
    \label{tab:asot}
\end{table}
    
\begin{table}
    \centering
    \setlength{\tabcolsep}{4.8pt}
    \renewcommand{\arraystretch}{1.3}  
    \resizebox{\columnwidth}{!}{%
    \begin{tabular}{|c|ccccc|ccccc|}
        \Xcline{1-11}{0.6pt}
        \multicolumn{11}{|c|}{\bf 50Salads Fine} \\
        \Xcline{1-11}{0.6pt}
        \multirow{2}{*}[-0.2em]{\bf Model} & \multicolumn{5}{c|}{\bf Known} 
        & \multicolumn{5}{c|}{\bf Unknown} \\[0.5ex]
        \cline{2-11}
        & {\it MoF} & {\it Edit} & \multicolumn{3}{c|}{\it F1@\{10,25,50\}} 
        & {\it MoF} & {\it Edit} & \multicolumn{3}{c|}{\it F1@\{10,25,50\}} \\[0.5ex]
        \hline
        \rowcolor{orange!15} Ours (GGSM + UASA)  &
        {\bf 69.4} & {\bf 62.9} & {\bf 69.6} & {\bf 68.5} & {\bf 62.4} &
        {\bf 37.2} & {\bf 40.1} & {\bf 45.1} & {\bf 41.5} & {\bf 27.1} \\
        UASA + TW-FINCH~\cite{sarfraz2021twfinch} & 
        60.8 & 48.1 & 49.9 & 48.9 & 44.2 &
        22.8 & 31.2 & 26.2 & 24.4 & 16.5 \\
        \hline
    \end{tabular}%
    }
    \caption{Comparison of the ability to distinguish actions within unknown segments between our approach and TW-FINCH~\cite{sarfraz2021twfinch}, a state-of-the-art unsupervised boundary detector method. Best results are in {\bf bold}.}
    \label{tab:finch}
\end{table}

\paragraph{Ablation on Loss}
In Table~\ref{tab:loss}, we evaluate the impact of the balancing parameter $\lambda$ in our loss function, which controls the contribution of the contrastive loss used to separate embeddings of unknown actions. We report the performance of our method using several values of $\lambda$ during training of the backbone TASB. Notably, the contrastive loss significantly improves the performance. 

\begin{table}
    \centering
    \setlength{\tabcolsep}{4.8pt}
    \renewcommand{\arraystretch}{1.3}  
    \resizebox{\columnwidth}{!}{%
    \begin{tabular}{|c|ccccc|ccccc|}
        \Xcline{1-11}{0.6pt}
        \multicolumn{11}{|c|}{\bf 50Salads Fine} \\
        \Xcline{1-11}{0.6pt}
        \multirow{2}{*}[-0.2em]{$\boldsymbol{\lambda}$} & \multicolumn{5}{c|}{\bf Known} 
        & \multicolumn{5}{c|}{\bf Unknown} \\[0.5ex]
        \cline{2-11}
        & {\it MoF} & {\it Edit} & \multicolumn{3}{c|}{\it F1@\{10,25,50\}} 
        & {\it MoF} & {\it Edit} & \multicolumn{3}{c|}{\it F1@\{10,25,50\}} \\[0.5ex]
        \hline
        $0$ & 
        64.9 & 53.4 & 60.1 & 58.2 & 52.2 &
        27.9 & 27.6 & 28.1 & 26.2 & 16.0 \\
        $0.01$ & 
        63.6 & 53.4 & 59.4 & 57.1 & 51.1 &
        29.4 & 30.8 & 32.4 & 29.2 & 18.6 \\
        \rowcolor{orange!15} $0.1$ &
        {\bf 69.4} & {\bf 62.9} & {\bf 69.6} & {\bf 68.5} & {\bf 62.4} &
        {\bf 37.2} & 40.1 & {\bf 45.1} & {\bf 41.5} & {\bf 27.1} \\
        $0.5$ & 
        66.6 & 55.3 & 64.4 & 63.2 & 56.8 &
        36.3 & {\bf 43.8} & 44.6 & 39.4 & 26.8 \\
        $1$ & 
        62.6 & 53.3 & 58.9 & 58.4 & 51.7 &
        33.3 & 43.3 & 43.2 & 36.0 & 21.3\\
        \hline
    \end{tabular}%
    }
    \caption{Impact of different values of the balancing parameter $\lambda$ in the loss function. Without contrastive loss ($\lambda=0$), the performance on unknown actions degrades. Best results are in {\bf bold}.}
    \label{tab:loss}
\end{table}

\begin{table}
    \centering
    \setlength{\tabcolsep}{4.8pt}
    \renewcommand{\arraystretch}{1.3}  
    \resizebox{\columnwidth}{!}{%
    \begin{tabular}{|c|ccccc|ccccc|}
        \Xcline{1-11}{0.6pt}
        \multicolumn{11}{|c|}{\bf 50Salads Fine} \\
        \Xcline{1-11}{0.6pt}
        \multirow{2}{*}[-0.2em]{$\boldsymbol{\alpha}$} & \multicolumn{5}{c|}{\bf Known} 
        & \multicolumn{5}{c|}{\bf Unknown} \\[0.5ex]
        \cline{2-11}
        & {\it MoF} & {\it Edit} & \multicolumn{3}{c|}{\it F1@\{10,25,50\}} 
        & {\it MoF} & {\it Edit} & \multicolumn{3}{c|}{\it F1@\{10,25,50\}} \\[0.5ex]
        \hline
        $0$ & 
        11.6 & 6.8 & 5.4 & 1.8 & 0.3 &
        12.8 & 7.6 & 13.1 & 7.9 & 0.0 \\
        $0.0001$ & 
        56.6 & 45.9 & 51.0 & 49.2 & 41.8 &
        33.7 & 36.4 & 40.4 & 36.1 & 22.7 \\
        \rowcolor{orange!15} $0.001$ &
        {\bf 69.4} & {\bf 62.9} & {\bf 69.6} & {\bf 68.5} & {\bf 62.4} &
        {\bf 37.2} & 40.1 & {\bf 45.1} & {\bf 41.5} & {\bf 27.1} \\
        $0.01$ & 
        66.6 & 58.8 & 66.2 & 64.5 & 58.1 &
        37.1 & {\bf 40.4} & 44.3 &  40.4 & 26.4 \\
        $0.1$ & 
        63.6 & 56.3 & 62.9 & 61.3 & 54.2 &
        34.6 & 38.8 & 42.3 & 36.9 & 24.4 \\
        $1$ & 
        57.6 & 50.3 & 54.9 & 53.0 & 44.5 &
        31.7 & 36.0 & 38.9 & 34.2 & 19.1 \\
        \hline
    \end{tabular}%
    }
    \caption{Impact of the parameter $\alpha$, which controls the balancing term in GGSM, on segmentation performance. Adding the balancing term ($\alpha>0$) improves the performance substantially.   
    Best results are in {\bf bold}.}
    \label{tab:alpha}
\end{table}

\paragraph{Ablation on $\boldsymbol{\alpha}$}
The parameter $\alpha$ steers the impact of the balancing term in GGSM, which prefers segments of the same length if two segments overlap but are not perfectly aligned. In Table \ref{tab:alpha}, we report the results for different values of $\alpha$. Setting $\alpha=0$ deactivates the balancing term, which leads to very low performance. 
As the tolerance increases, performance improves, reaching an optimal value at $\alpha=0.001$. However, further increasing $\alpha$ gives higher weight to the same length than to the alignment, which degrades performance for $\alpha>0.1$. 

\paragraph{Ablation on UASA without Segments}
\label{sec:alpha}
The initialization of unknown clusters in Unknown Action Segment Assignment (UASA) is performed using the mean value of the unknown segments identified by GGSM. This strategy ensures a more stable and representative initialization of the clusters. Without leveraging segment-based  initialization, centroids would have to be determined from individual frame embeddings, which are inherently noisier. This increased noise makes the estimation of the number of unknown actions $K$ less reliable, as the Gaussian Mixture Model struggles with the higher variability in frame-level embeddings. 
We show this behavior in Table \ref{tab:segm-init}.

\begin{table}[tbp]
    \centering
    \setlength{\tabcolsep}{4.8pt}
    \renewcommand{\arraystretch}{1.3}  
    \resizebox{.47\textwidth}{!}{%
    \begin{tabular}{|c|ccccc|ccccc|}
        \Xcline{1-11}{0.6pt}
        \multicolumn{11}{|c|}{\bf 50Salads Fine} \\
        \Xcline{1-11}{0.6pt}
        {\bf Segment} & \multicolumn{5}{c|}{\bf Known} & \multicolumn{5}{c|}{\bf Unknown} \\[0.5ex]
        \cline{2-11}
        {\bf Init.} & {\it MoF} & {\it Edit} & \multicolumn{3}{c|}{\it F1@\{10,25,50\}} 
        & {\it MoF} & {\it Edit} & \multicolumn{3}{c|}{\it F1@\{10,25,50\}} \\[0.5ex]
        \hline
        \rowcolor{orange!15} \ding{51} &
        {\bf 69.4} & 62.9 & {\bf 69.6} & {\bf 68.5} & {\bf 62.4} &
        {\bf 37.2} & 40.1 & {\bf 45.1} & {\bf 41.5} & {\bf 27.1} \\
        \ding{55} & 
        68.5 & {\bf 63.0} & 69.4 & 67.8 & 60.7 &
        31.8 & {\bf 41.6} & 36.5 & 32.8 & 22.8 \\
        \hline
    \end{tabular}%
    }
    \caption{Impact of different initialization strategies for unknown clusters in the Unknown Action Segment Assignment (UASA) module: segment-based initialization against frame-based initialization. Segment-based initialization provides more robust representations and improves the estimation of the number of unknown actions $K$, leading to better overall performance. Best results are in {\bf bold}}
    \label{tab:segm-init}
\end{table}

\begin{figure}[tbp]
    \centering
    \includegraphics[width=\columnwidth]{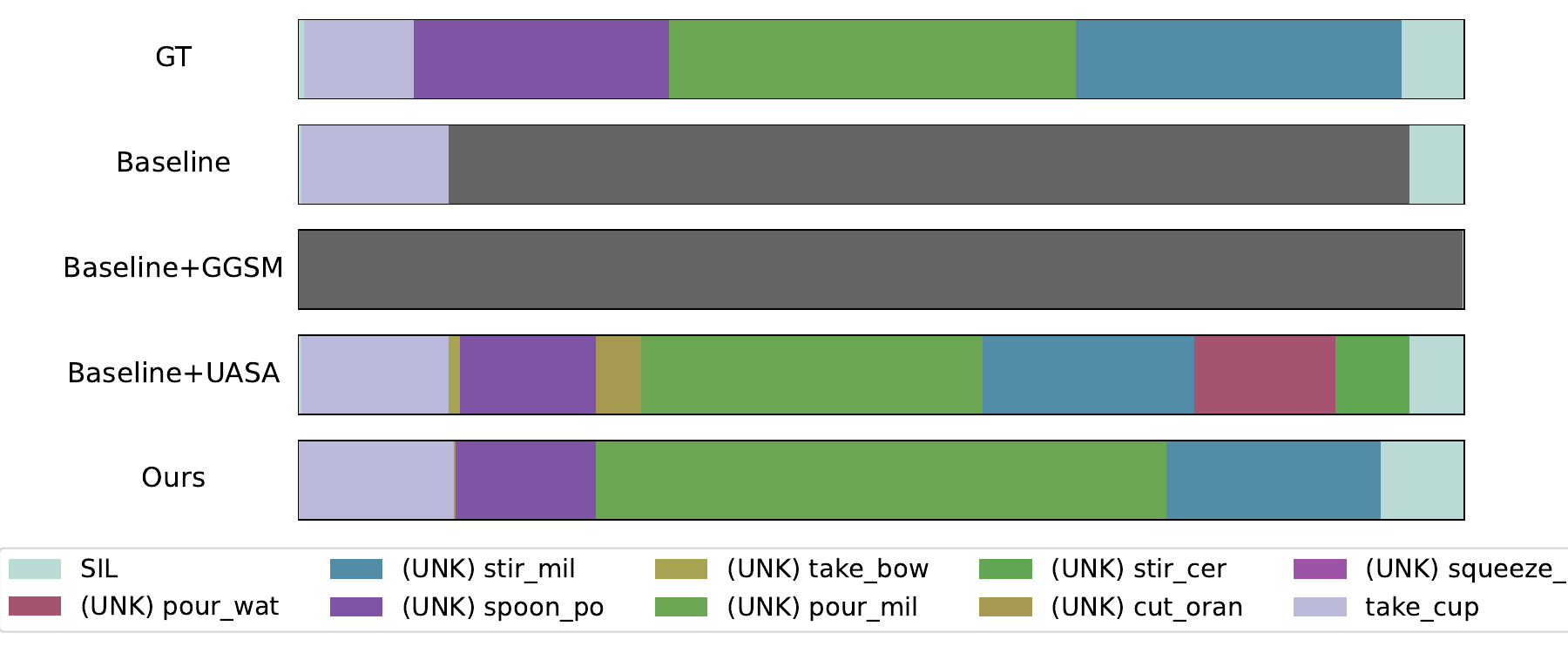}
    \hfill
    \centering
    \captionsetup{type=figure}
    \caption{Qualitative results on Breakfast dataset, {\it "P20\_stereo01\_P20\_milk"}. The color gray represents the label {\tt UNK} predicted by TASB. Faded colors represent known actions.}
    \label{fig:breakfast-qualitative}
\end{figure}

\paragraph{Qualitative Results}
Figure \ref{fig:breakfast-qualitative} shows qualitative results for a video from the Breakfast dataset. The purple, green and blue intense colors are the unknown actions, corresponding to {\it spoon\_powder}, {\it pour\_milk} and {\it stir\_milk}. Baseline shows that the TASB is able to effectively discriminate between unknown and known actions. However, in Baseline+GGSM, the boundaries are not clearly visible, as both unknown actions are assigned the same label. The combination of Granularity-Guided Segmentation Module (GGSM) and the Unknown Action Segment Assignment (UASA) yields a good segmentation of the three unknown actions.  

\section{Conclusions}

In this work, we introduced {\it Action Discovery}, a new setup within Temporal Action Segmentation that addresses the challenges of partially annotated data, where only a subset of actions - {\it known} actions - are labeled, while other {\it unknown} actions remain unlabeled. 
This protocol enables the segmentation and identification of ambiguous or overlooked behaviors with the same granularity as the labeled actions.
To tackle this problem, we proposed two key components: the Granularity-Guided Segmentation Module (GGSM) for accurate boundary detection within unlabeled segments, and the Unknown Action Segment Assignment (UASA), a similarity-based method that assigns plausible labels to unknown segments without requiring the knowledge of the number of unknown actions.
Through extensive experiments on Breakfast, 50Salads and DesktopAssembly, we demonstrated that our approach effectively captures the structure of unknown actions and leverages the granularity information of known actions to improve segmentation accuracy. 
These findings highlight the potential of Action Discovery to support the refinement of partially annotated datasets and uncover meaningful behaviors, with promising applications in domains like neuroscience and behavior analysis.

\bibliographystyle{plainnat}
\bibliography{actiondiscovery}

\clearpage

\appendix
\twocolumn[{
\begin{center}
    \LARGE \textbf{Looking into the Unknown: Exploring Action Discovery for Segmentation of Known and Unknown Actions - Supplementary Materials} \\[1ex]
\end{center}
\vspace{2em}
}]

\setcounter{page}{1}
\setcounter{figure}{3}
\setcounter{table}{7} 

\section{Hardware}
All the experiments were performed on Ubuntu 24.04.2LTS, with 128GB of RAM, Intel i9-13900K CPU and a Nvidia RTX 4090 GPU. The software used is Python version 3.12.3, with PyTorch 2.3.0, Scikit-Learn 1.5.0 and Numpy 1.26.4. An environment file will be released along with the code, upon acceptance.

\section{Ablation}
In Table~\ref{tab:tau} we evaluate the impact of the temperature parameter $\tau$ in the contrastive loss $\mathcal{L}_{contr}$. When $\tau$ is high (\eg $\tau > 0.5$), the effect of the contrastive loss becomes negligible due to poor separation between positive and negative samples. Conversely, with a low value of $\tau$, the softmax becomes very sharp, and even small differences in similarity can dominate the loss. An intermediate value, \ie $\tau=0.4$, yields the best performances across most metrics, for both known and unknown evaluations.  
\begin{table}[tbp]
    \centering
    \setlength{\tabcolsep}{4.8pt}
    \renewcommand{\arraystretch}{1.3}  
    \resizebox{.47\textwidth}{!}{%
        \begin{tabular}{|c|ccccc|ccccc|}
            \Xcline{1-11}{0.6pt}
            \multicolumn{11}{|c|}{\bf 50Salads Fine} \\
            \Xcline{1-11}{0.6pt}
            \multirow{2}{*}[0em]{$\boldsymbol{\tau}$} & \multicolumn{5}{c|}{\bf Known} & \multicolumn{5}{c|}{\bf Unknown} \\[0.5ex]
            \cline{2-11}
            & {\it MoF} & {\it Edit} & \multicolumn{3}{c|}{\it F1@\{10,25,50\}} 
            & {\it MoF} & {\it Edit} & \multicolumn{3}{c|}{\it F1@\{10,25,50\}} \\[0.5ex]
            \hline
            0.1 &
            67.5 & 58.3 & 65.6 & 62.8 & 57.1 &
            37.0 & 37.3 & 42.6 & 39.5 & \textbf{27.8} \\
            0.2 &
            \textbf{71.1} & 62.5 & 69.4 & 66.7 & 61.0 &
            34.5 & 39.0 & 41.7 & 39.6 & 24.1 \\
            \rowcolor{orange!15} 0.4 &
            69.4 & 62.9 & {\bf 69.6} & {\bf 68.5} & {\bf 62.4} &
            {\bf 37.2} & \textbf{40.1} & {\bf 45.1} & {\bf 41.5} & 27.1 \\
            0.6 &
            70.3 & \textbf{63.5} & 67.1 & 65.2 & 59.2 &
            32.8 & 33.6 & 39.2 & 35.9 & 24.7 \\
            0.8 & 
            70.1 & 62.0 & 67.2 & 65.2 & 59.3 &
            31.3 & 35.3 & 37.0 & 34.9 & 22.4 \\
            1.0 &
            70.5 & 61.8 & 67.5 & 64.1 & 59.7 &
            27.8 & 34.0 & 32.9 & 29.4 & 18.5 \\
            \hline
        \end{tabular}%
    }
    \caption{Impact of different temperature values $\tau$ on the contrastive loss $\mathcal{L}_{contr}$. We use $\tau=0.4$. Best results are in {\bf bold}}
    \label{tab:tau}
\end{table}

\section{Inference Time}
The training and inference time of our approach largely depend on the backbone used in TASB. While our modules do not affect the training time, they can impact inference time. Specifically, UASA uses the training data to initialize the number $K$ of unknown actions and the cluster centroids, whereas GGSM employs hierarchical clustering to compute the final temporal spans. Table~\ref{tab:times} shows the inference time impact of our modules when using MS-TCN as the backbone and one split of the 50 Salads Fine dataset. GGSM introduces only a minor overhead, and the complete method (GGSM+UASA) is slower than the baseline alone, but remains around one minute. Notably, UASA alone - without segment initialization - must process every unknown embeddings and its runtime depends heavily on the search range of $K$. Consequently, UASA without GGSM, which allows segment initialization, takes approximately $20$ minutes.

\begin{table}[tbp]
    \centering
    \resizebox{.25\textwidth}{!}{%
    \begin{tabular}{|cc|c|}
        \hline
        \multicolumn{3}{|c|}{\bf 50Salads Fine} \\
        \hline
        GGSM & UASA & Time (s) \\ 
        \hline
        & & \textbf{9.87} \\
        \ding{51} & & 21.2 \\
        & \ding{51} & 1169.2 \\
        \rowcolor{orange!15} \ding{51} & \ding{51} & 83.0 \\
        \hline
    \end{tabular}%
    }
    \caption{Impact of individual modules on inference time. Best results are in {\bf bold}}
    \label{tab:times}
\end{table}

\section{Qualitative Results}
\label{sec:qualitative}
\begin{figure*}[htbp]
    \centering
    \includegraphics[width=0.9\linewidth]{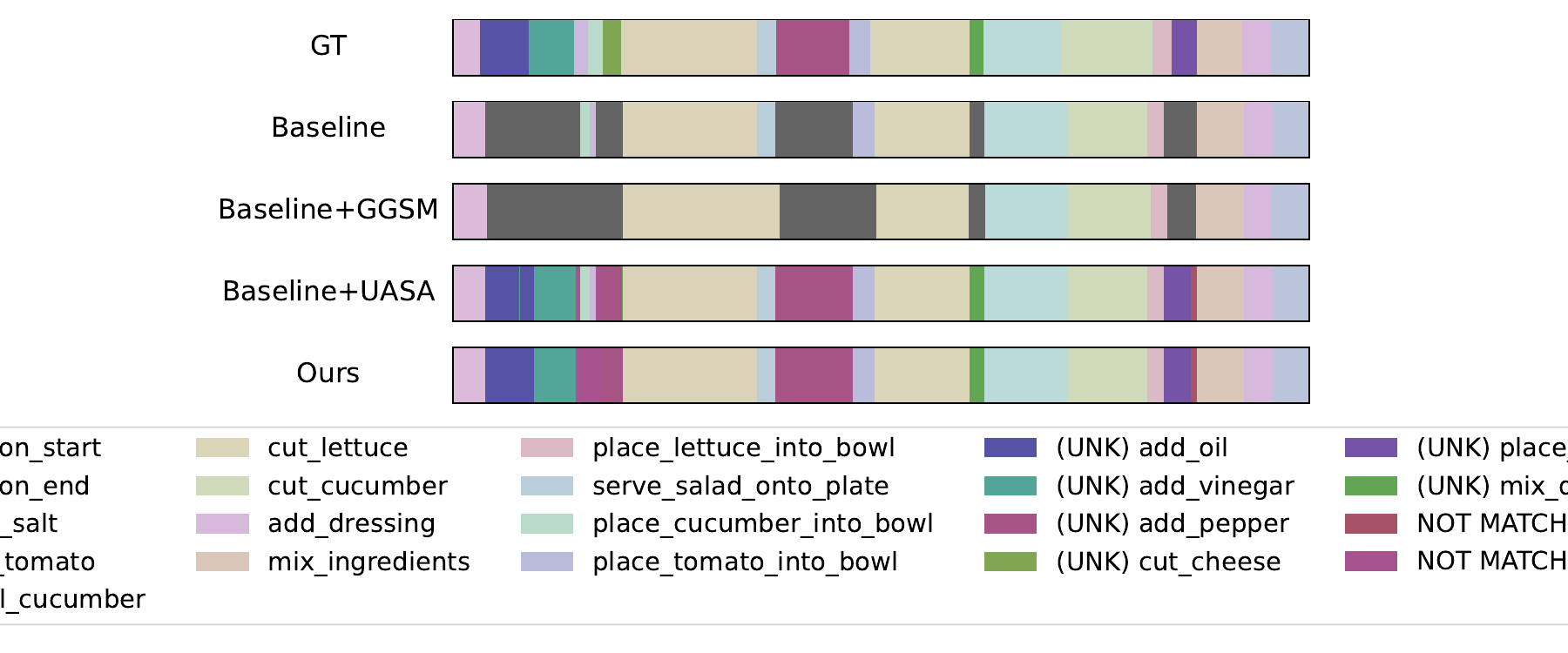}
    \caption{Qualitative results for the video {\it rgb-13-2} of 50Salads dataset, at Fine granularity. More faded colors represent the known actions, less faded segments represent the unknown actions.}
    \label{fig:fsf-qual}
\end{figure*}

\begin{figure*}[htbp]
    \centering
    \includegraphics[width=0.9\linewidth]{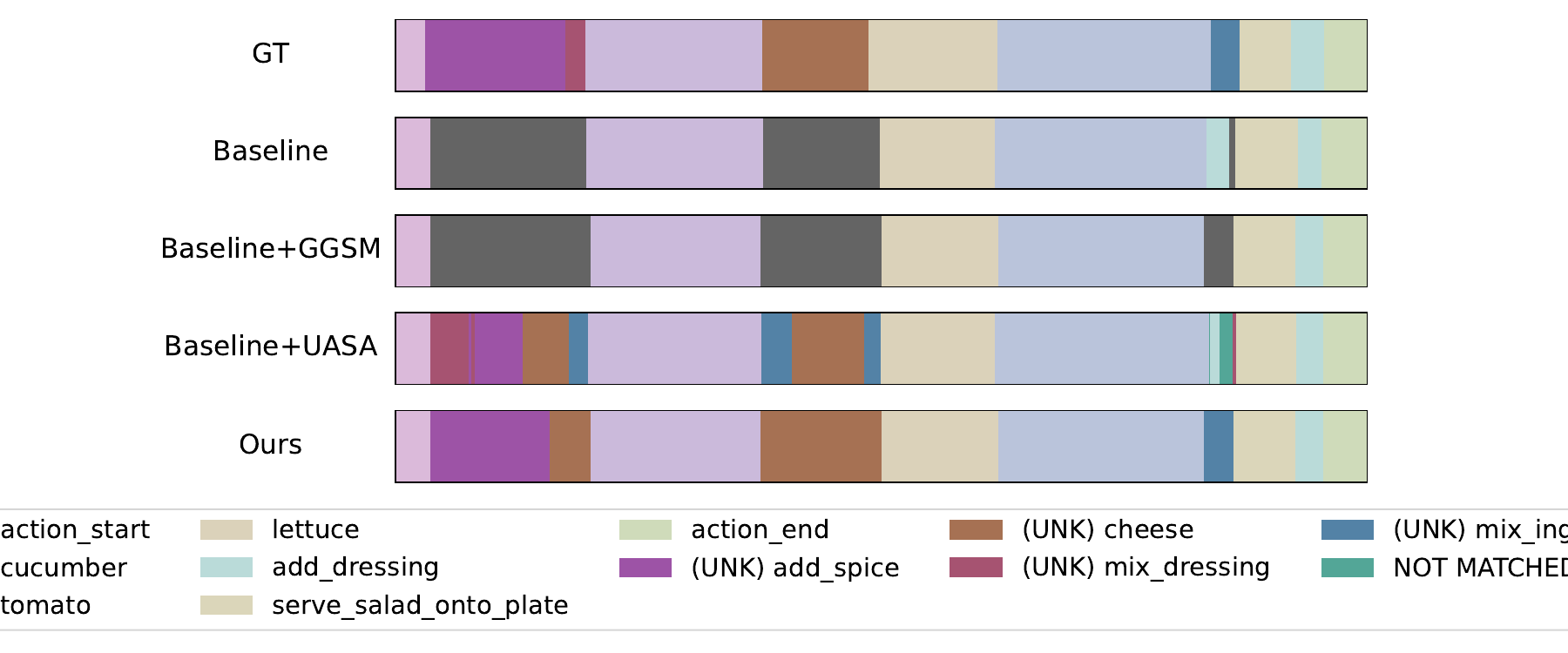}
    \caption{Qualitative results for the video {\it rgb-13-2} of 50Salads dataset, at Coarse granularity. More faded colors represent the known actions, less faded segments represent the unknown actions.}
    \label{fig:fsc-qual}
\end{figure*}

Figure \ref{fig:fsf-qual} and Figure \ref{fig:fsc-qual} illustrate qualitative results for the same video from the 50Salads dataset at Fine and Coarse granularity. Our model demonstrates precise segmentation for both known and unknown actions. Notably, our method successfully segments the first unknown segment based on the relative granularity: it detects four actions for fine-granularity and two for coarse-granularity. Furthermore, the UASA module effectively assigns the majority of the unknown segments. In contrast, Baseline and Baseline+GGSM are not designed for differentiating between unknown actions.

\begin{figure*}[tbp]
    \centering
    \includegraphics[width=0.9\linewidth]{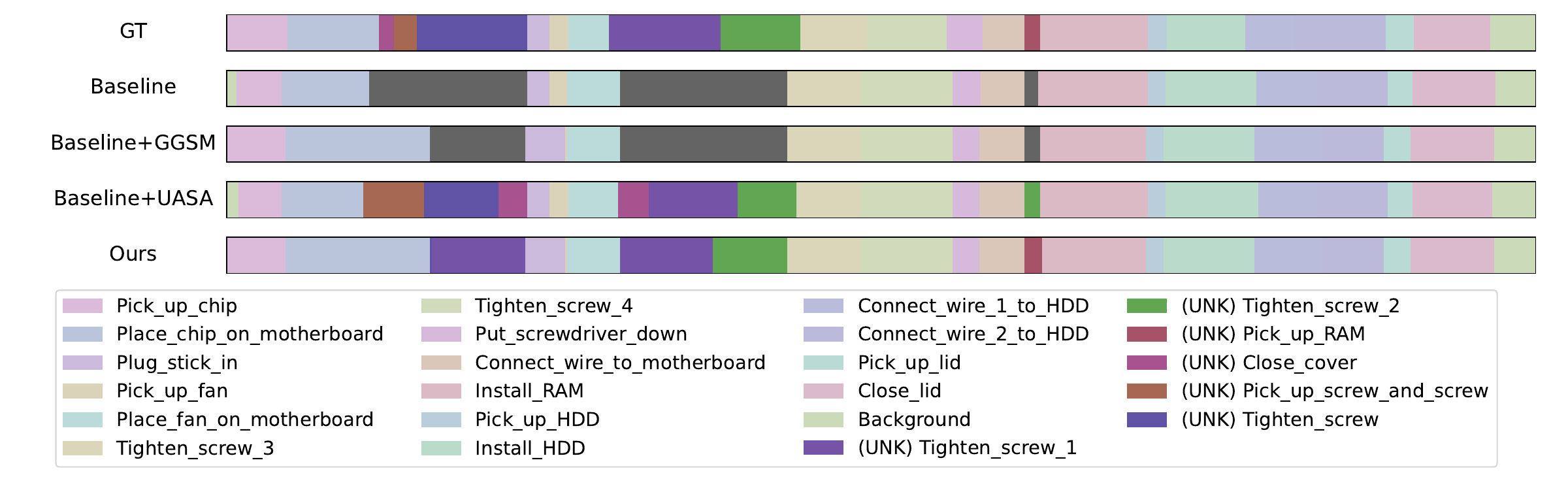}
    \caption{Qualitative results for the video {\it 2020-04-19\_17-20-50} of DesktopAssembly dataset. More faded colors represent the known actions, less faded colors represent the unknown actions.}
    \label{fig:da-qual}
\end{figure*}

Qualitative results for the DesktopAssembly dataset are shown in Figure \ref{fig:da-qual}. Our method provides accurate segmentation and action assignment for the second unknown segment, and it also effectively segments the third, very short unknown action segments, while still maintaining excellent segmentation of known actions.

\section{List of Known Actions}
\label{sec:known}
In Table \ref{tab:known-list} we report the list of known actions for each dataset considered in this work. In Table \ref{tab:coarse-fine}, we report how we transition from fine-grained annotations of 50Salads to coarse-grained ones by merging the fine actions.

\begin{table*}[tbp]
    \centering
    \begin{tabular}{|c|c|}
        \hline
        \multicolumn{2}{|c|}{\bf 50Salads} \\
        \hline
        {\bf Coarse Action} & {\bf Fine Action} \\
        \hline
        action\_start & action\_start \\
        \hline
        action\_end & action\_end \\
        \hline
        add\_dressing & add\_dressing \\
        \hline
        mix\_dressing & mix\_dressing \\
        \hline
        mix\_ingredients & mix\_ingredients \\
        \hline
        serve\_salad\_onto\_plate & serve\_salad\_onto\_plate \\
        \hline        
        \multirow{2}{*}[0em]{lettuce} & cut\_lettuce \\
        & place\_lettuce\_into\_bowl \\
        \hline        
        \multirow{3}{*}[0em]{cucumber} & peel\_cucumber  \\
         & cut\_cucumber \\
         & place\_cucumber\_into\_bowl \\
         \hline
         \multirow{2}{*}[0em]{tomato} & cut\_tomato \\
         & place\_tomato\_into\_bowl \\
         \hline
         \multirow{4}{*}[0em]{add\_spice} & add\_vinegar \\
         & add\_pepper \\
         & add\_salt \\
         & add\_oil \\
         \hline
         \multirow{2}{*}[0em]{cheese} & cut\_cheese \\
         & place\_cheese\_into\_bowl \\
         \hline
    \end{tabular}
    \caption{On the left, the coarse granularity annotations of 50Salads; on the right, the finer-grained actions that compose each coarse action.}
    \label{tab:coarse-fine}
\end{table*}

\begin{table*}[tb]
    \centering
    \resizebox{1.\textwidth}{!}{%
    \begin{tabular}{|c|c|c|c|}
        \hline
        \rowcolor{orange!15} \multicolumn{4}{|c|}{\bf Unknown Actions} \\
        \hline
        {\bf Breakfast}&{\bf 50Salads Fine}&{\bf 50Salads Coarse}&{\bf DesktopAssembly}\\ 
        \hline
        pour\_sugar & add\_oil & add\_spice & Tighten\_screw\_1 \\
        pour\_water & add\_vinegar & cheese & Tighten\_screw\_2 \\
        add\_teabag & cut\_cheese & mix\_dressing & Pick\_up\_RAM \\
        stir\_milk & mix\_dressing & mix\_ingredients & Close\_cover \\
        spoon\_powder & mix\_ingredients & - & Pick\_up\_screw\_and\_screwdriver \\
        pour\_oil & place\_lettuce\_into\_bowl & - & Tighten\_screw \\
        stir\_egg & - & - & - \\
        butter\_pan & - & - & - \\
        take\_bowl & - & - & - \\
        pour\_milk & - & - & - \\
        stir\_cereals & - & - & - \\
        crack\_egg & - & - & - \\
        take\_knife & - & - & - \\
        cut\_orange & - & - & - \\
        squeeze\_orange & - & - & - \\
        
        \hline
    \end{tabular}
    }
    \caption{List of unknown actions for Breakfast, 50Salads Fine, Coarse and DesktopAssembly.}
    \label{tab:unknown-list}
\end{table*}

\begin{table*}[tb]
    \centering
    \resizebox{1.\textwidth}{!}{%
    \begin{tabular}{|c|c|c|c|}
        \hline
        \rowcolor{orange!15} \multicolumn{4}{|c|}{\bf Known Actions} \\
        \hline
        {\bf Breakfast}&{\bf 50Salads Fine}&{\bf 50Salads Coarse}&{\bf DesktopAssembly}\\ 
        \hline
        SIL&action\_start&action\_start&Pick\_up\_chip\\
        put\_toppingOnTop&action\_end&cucumber&Place\_chip\_on\_motherboard\\
        put\_egg2plate&add\_salt&tomato&Plug\_stick\_in\\
        pour\_cereals&cut\_tomato&lettuce&Pick\_up\_fan\\
        smear\_butter&peel\_cucumber&add\_dressing&Place\_fan\_on\_motherboard\\
        spoon\_sugar&cut\_lettuce&serve\_salad\_onto\_plate&Tighten\_screw\_3\\
        pour\_flour&cut\_cucumber&action\_end&Tighten\_screw\_4\\
        fry\_pancake&add\_dressing&-&Put\_screwdriver\_down\\
        put\_fruit2bowl&serve\_salad\_onto\_plate&-&Connect\_wire\_to\_motherboard\\
        take\_eggs&place\_cucumber\_into\_bowl&-&Install\_RAM\\
        put\_bunTogether&place\_tomato\_into\_bowl&-&Pick\_up\_HDD\\
        stir\_dough&add\_pepper&-&Install\_HDD\\
        add\_saltnpepper&place\_cheese\_into\_bowl&-&Connect\_wire\_1\_to\_HDD\\
        cut\_bun&-&-&Connect\_wire\_2\_to\_HDD\\
        pour\_dough2pan&-&-&Pick\_up\_lid\\
        take\_topping&-&-&Close\_lid\\
        stir\_fruit&-&-&Background\\
        take\_glass&-&-&-\\
        fry\_egg&-&-&-\\
        peel\_fruit&-&-&-\\
        stirfry\_egg&-&-&-\\
        pour\_egg2pan&-&-&-\\
        put\_pancake2plate&-&-&-\\
        take\_cup&-&-&-\\
        take\_plate&-&-&-\\
        stir\_coffee&-&-&-\\
        pour\_coffee&-&-&-\\
        spoon\_flour&-&-&-\\
        take\_squeezer&-&-&-\\
        stir\_tea&-&-&-\\
        take\_butter&-&-&-\\
        cut\_fruit&-&-&-\\
        pour\_juice&-&-&-\\
        \hline
    \end{tabular}
    }
    \caption{List of known actions for Breakfast, 50Salads Fine, Coarse and DesktopAssembly.}
    \label{tab:known-list}
\end{table*}

\end{document}